\documentclass{article}

\PassOptionsToPackage{numbers, compress}{natbib}

\usepackage[final]{neurips_2022}

\title{Memory in humans and deep language models: Linking hypotheses for model augmentation}

\usepackage[utf8]{inputenc} 
\usepackage[T1]{fontenc}    
\usepackage{hyperref}       
\usepackage{url}            
\usepackage{booktabs}       
\usepackage{amsfonts}       
\usepackage{nicefrac}       
\usepackage{microtype}      
\usepackage{xcolor}         
\usepackage{graphicx}
\usepackage{wrapfig,floatrow}
\usepackage{lipsum}

\usepackage{soul}           

\author{%
  Omri Raccah \\
  Intel Labs \\
  New York University\\
  \texttt{or409@nyu.edu} \\
  \And
  Phoebe Chen \\
  New York University\\
  \texttt{hc2896@nyu.edu} \\
  \And
  Ted L. Willke \\
  Intel Labs \\
  \texttt{ted.willke@intel.com}
  \And
  David Poeppel \\
  New York University \\
  \texttt{dp101@nyu.edu} \\
  \And
  Vy A. Vo \\
  Intel Labs \\
  \texttt{vy.vo@intel.com} \\
}

\begin{document}

\maketitle

\begin{abstract}
  The computational complexity of the self-attention mechanism in Transformer models significantly limits their ability to generalize over long temporal durations. Memory-augmentation, or the explicit storing of past information in external memory for subsequent predictions, has become a constructive avenue for mitigating this limitation. We argue that memory-augmented Transformers can benefit substantially from considering insights from the memory literature in humans. We detail an approach for integrating evidence from the human memory system through the specification of cross-domain linking hypotheses. We then provide an empirical demonstration to evaluate the use of surprisal as a linking hypothesis, and further identify the limitations of this approach to inform future research. 
\end{abstract}

\section{Introduction}

Transformer model architectures \cite{vaswaniAttentionAllYou2017} have become an indispensable tool in several domains, such as natural language processing \cite{wolf2019huggingface}, image processing \cite{khanTransformersVisionSurvey2022} and reinforcement learning \cite{luketinaSurveyReinforcementLearning2019, meloTransformersAreMetaReinforcement2022}. A widely acknowledged scaling limitation of the self-attention mechanism is that its computational complexity scales quadratically with the size of the attention window. This limits the model's ability to capture long-range dependencies in data such as books, scientific articles, or code. Several efficient Transformers have been proposed to address this principal limitation \cite{tayEfficientTransformersSurvey2020}. A subset of these focus on augmenting the network with an external memory, which we henceforth refer to as memory-augmented Transformers \cite{raeCompressiveTransformersLongRange2019, guuREALMRetrievalAugmentedLanguage2020, fanAugmentingTransformersKNNBased2021, yogatamaAdaptiveSemiparametricLanguage2021}. Notably, knowledge in artificial neural networks (ANNs) is thought to be implicitly stored in the parameters of a pre-trained model, requiring ever-larger networks to store more facts \cite{nematzadehMemoryHumanArtificial2020,stephenson_geometry_2021}. In memory-augmented Transformers, however, information is more explicitly stored in an external memory and retrieved when making predictions. This may increase the capacity of the network to represent knowledge about the world, in addition to helping it capture information over long temporal durations. Unlike temporal convolutions, which require a pre-specified width, external memories can be stored for arbitrary durations and retrieved when relevant.

This property is also true of human memory, which demonstrates the remarkable ability to generalize over an immense amount of information in written documents and over events in one’s life. 
The rich literature on memory in psychology and neuroscience presents ample opportunities for augmenting ANNs with biologically-inspired memory. 
Here, we aim to lay some groundwork for understanding memory across fields, and describe some practical considerations for effectively integrating findings from cognitive neuroscience into Transformers, with a particular focus on language models (LMs). 
However, note that several of these considerations can be applied to other models and domains in AI. Finally, we provide an empirical demonstration of evaluating a memory augmentation strategy for GPT-2 \cite{radfordLanguageModelsAre} using human behavioral data and identify its limitations and strengths to inform future research.

\section{Considerations for biologically-inspired memory augmentation}

\subsection{Memory-augmentation from a cognitive lens}
We argue that specifying appropriate linking hypotheses across domains will not only facilitate novel biologically-inspired approaches, but will also provide a way to empirically evaluate different hypotheses.
In cognitive neuroscience, a linking hypothesis is a proposition for a formal mapping between a neurobiological state and a psychological state \cite{tellerLinkingPropositions1984, poeppelMapsProblemMapping2012}, such as the firing of a single neuron leading to a visual percept. 
A central aim of biologically-inspired AI is to formulate linking hypotheses between a component in an AI system and a well-defined aspect of cognition. 
Strong linking hypotheses should lead to a formal and quantifiable mapping between a representation in an AI system and some neurobiological/psychological data, as has been demonstrated in some cases for computer vision \cite{cichy_comparison_2016,kubilius_brain-like_2019,yamins_using_2016} and natural language \cite{heilbron_hierarchy_2022,schrimpf_neural_2021,wilcox_predictive_2020,wehbe_simultaneously_2014}. These linking hypotheses must be specified at the correct level of analysis \cite{poeppelMapsProblemMapping2012, marrVisionComputationalInvestigation1982}, e.g., a modification of the equations to perform similarity search on a database in a retrieval-augmented system should map to research on the biological mechanisms of memory retrieval. 
In our view, proper accounts would be best derived through decomposing the problem into computational subroutines appropriate for comparison across domains. Many AI systems already assume a linking hypothesis between ANNs and human cognition without explicitly stating them as hypotheses or evaluating them. Here, we briefly explore some of these hypotheses in memory-augmented Transformers and propose possible mappings to findings in the human literature.

We divide memory-augmented Transformers into two general types. 
A \emph{static} memory stores information in a corpus of fixed size and content (e.g., a Wikipedia knowledge base), which it learns to retrieve from during training \cite{khandelwalGeneralizationMemorizationNearest2020, lewisRetrievalAugmentedGenerationKnowledgeIntensive2020, fanAugmentingTransformersKNNBased2021}. The contents of a static memory do not get modified, although they can be encoded in different formats such as raw text or embeddings \cite{lewisRetrievalAugmentedGenerationKnowledgeIntensive2020, fevryEntitiesExpertsSparse2020}.
\emph{Dynamic} memory mechanisms store new information as inputs that are being processed by the model. Training the network involves learning both storage and retrieval policies. 
For example, new information may be remembered or forgotten on the basis of 
input properties or model activations. 
Furthermore, inputs may be transformed in some manner (e.g., through compression) before being stored in the external memory \cite{raeCompressiveTransformersLongRange2019}. Both static and dynamic memory-augmented Transformers have shown significant improvements over non-augmented models when making predictions over long texts \cite{raeCompressiveTransformersLongRange2019, guuREALMRetrievalAugmentedLanguage2020, daiTransformerXLAttentiveLanguage2019, wuMemorizingTransformers2022}.

These augmentation strategies do not map cleanly to the types of memory commonly delineated in cognitive theories of human memory \cite{tulvingEpisodicSemanticMemory1972}. That said, classical memory taxonomies are often the source of AI inspiration, with papers citing work on short- vs. long-term memory or episodic memory \cite{gravesNeuralTuringMachines2014, gravesHybridComputingUsing2016, yogatamaAdaptiveSemiparametricLanguage2021, nematzadehMemoryHumanArtificial2020}. In our view, a static memory could be like human semantic memory if it uses a knowledge base, or it could be a fairly direct analog of episodic memory if it stores previously seen examples \cite{khandelwalGeneralizationMemorizationNearest2020}. Instead, our proposed division focuses on the subprocesses thought to be involved in human long-term memory: encoding, consolidation, and retrieval \cite{purves_textbook}. Different strategies for memory augmentation will therefore pursue different implementations of each subprocess, and can draw direct inspiration from studies of that specific subprocess. Current work on memory-augmented Transformers has already proposed separate mechanisms for each subprocess, although there is often no direct link to human data. For example, there is a growing literature on retrieval augmentation \cite{guuREALMRetrievalAugmentedLanguage2020, fanAugmentingTransformersKNNBased2021, khandelwalGeneralizationMemorizationNearest2020, lewisRetrievalAugmentedGenerationKnowledgeIntensive2020, blattmannRetrievalAugmentedDiffusionModels2022, borgeaudImprovingLanguageModels2022, ashualKNNDiffusionImageGeneration2022} that proposes similarity search as the retrieval mechanism. Other work has proposed specific encoding policies which determine what to store and forget, either by exploiting the attention weights \cite{raeCompressiveTransformersLongRange2019} or learning which memories to forget \cite{sukhbaatarNotAllMemories2021}.


\subsection{Incorporating insights from human memory via policy modifications} 

Here we 
discuss some findings from the
human memory literature to demonstrate how they may be used to inform policy modifications in memory-augmented Transformers.  
Lexical properties (e.g., written-frequency, word length, animacy, etc.) serve as strong predictors of subsequent memory for individual words and lists \cite{hulme_word-frequency_1997,popp_animacy_2018,aka_predicting_2021}. Furthermore, humans have been shown to have the remarkable ability to remember whether they have seen an image from up to 10,000 images after only a single exposure \cite{standingLearning10000Pictures1973}. The properties that determine the memorability of an image are thought to be multifaceted, including high-level properties such as emotional valence \cite{khoslaUnderstandingPredictingImage2015} and overall semantic meaning \cite{bainbridgeMemorabilityHowWhat2019, rustUnderstandingImageMemorability2020}.
If some property is directly computable from the inputs, it can be efficiently used as a biologically-plausible \emph{encoding policy} in memory-augmented models.
Recent work in cognitive neuroscience has also been focused on uncovering the process by which humans segment continuous experience into composite events in memory, known as event segmentation \cite{zacks2001event,wangUncoveringSurprisingEvent2022,raccah2022acoustic,kumar_goldstein_michelmann_zacks_hasson_norman_2022}. This evidence can also inform encoding policies for model augmentation, as studies have shown preferential encoding at event boundaries. Furthermore, this area of research can be leveraged to inform \emph{storage policies}, which delineate how sequential information with ordered constituents is structured or formatted in memory. Lastly, \emph{retrieval policies}, or the manner by which information is read from an existing memory store, can take practical influence from human memory. For example, items that share a temporal or semantic context during encoding are retrieved sequentially with relation to one another \cite{howard2002distributed,manning2012interpreting}. 
These examples provide theoretically and empirically motivated hypotheses for memory-augmentation. Next, we demonstrate the evaluation of a specific linking hypothesis.

\section{Evaluating a candidate linking hypothesis for memory augmentation} 
\label{sec:empirical}

\paragraph{Surprisal}
The loss function of an LM estimates the negative log likelihood of an upcoming word given its context. In information theoretic terms, this is known as \textit{surprisal}. Some have proposed that next-word prediction is a fundamental computational process that occurs during human language processing \cite{schrimpfNeuralArchitectureLanguage2021, goldsteinSharedComputationalPrinciples2022, shainFMRIRevealsLanguagespecific2020}, and have shown evidence that LM-estimated surprise predicts behavioral \cite{goodkindPredictivePowerWord2018, hao2020, jacobs-mccarthy-2020-human} and neural data \cite{goldsteinSharedComputationalPrinciples2022}. Surprise (or unsigned prediction error) is also theorized to play a critical role in memory and learning, and experimental evidence supports this notion \cite{sinclairSurpriseDestabilizePrediction2018, fosterRoleSurpriseLearning2019, antonyBehavioralPhysiologicalNeural2020}. 
Word surprisal in particular may predict human memory during natural language comprehension \cite{futrellLossyContextSurprisalInformationTheoretic2020, haeuserEffectsPredictionError2021}.
Since surprise is a readily available quantity in LMs, we test its feasibility as a linking hypothesis by examining human behavioral data in a memory experiment. If model-based surprisal can predict human memory, it could be a practical and effective memory encoding policy for augmented models.

\paragraph{Dataset of human recall behavior}
We used a public
dataset collected by Michelmann et al. from two groups of participants \cite{michelmannMomentbymomentTrackingNaturalistic2021}. The first group ("story-exposure"; N = 50) listened to a naturally spoken story containing 965 words. Then participants completed a cloze task \cite{taylorClozeProcedureNew1953} similar to an autoregressive 
LM objective, in which they were given 10 words from the story and asked to predict the final word. This task was administered in order for every word in the story, 
starting with the third word, limiting the context for words in the beginning of the story (Appendix \ref{methods}). 
The second group ("no exposure"; N = 50) completed the same cloze task but had no exposure to the story before completing the cloze task. 
The memory effect is
the difference in performance across groups.
For a full account of the methods, see Appendix \ref{methods}.

\begin{wrapfigure}[16]{R}{0.6\textwidth}
    \centering
    \includegraphics[width=.8\textwidth, trim={3.5cm 1cm 4cm 3cm}]{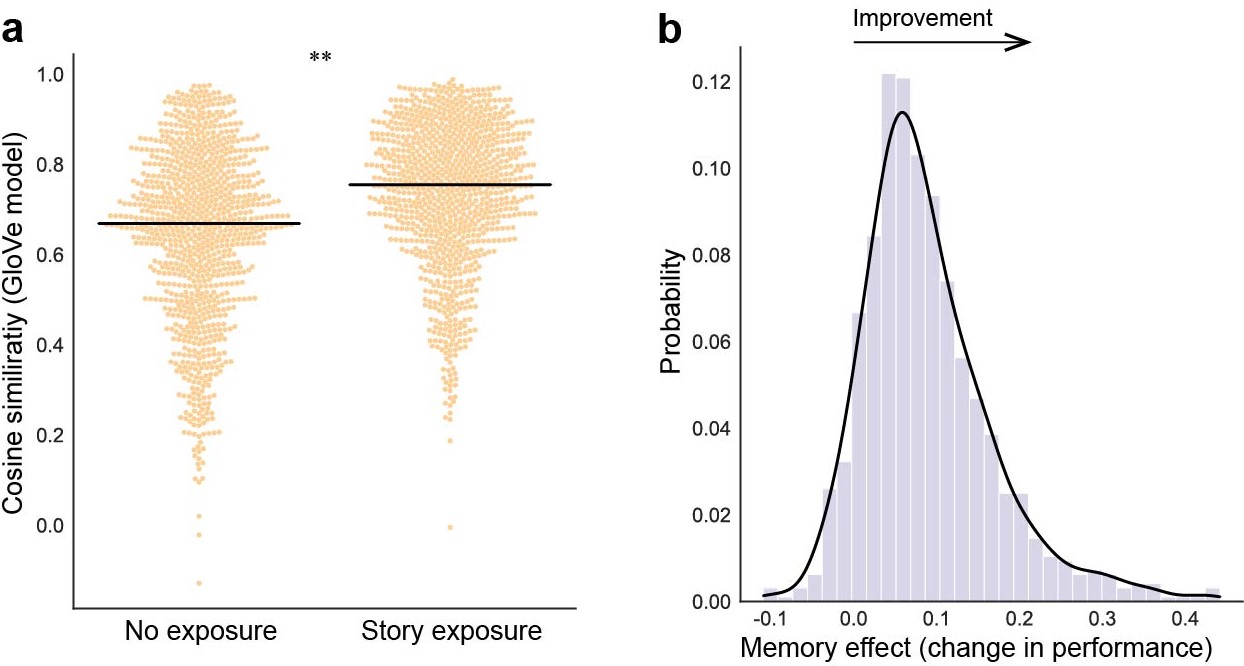}
    \captionsetup{width=1\textwidth}
    \caption{Behavioral results. (A) Cloze performance increases as a function of story-exposure across individual words. Black lines indicate the mean. (B) Histogram of memory improvement across words (signed difference between the story-exposure and no-exposure groups.)}
    \label{fig:histograms}
\end{wrapfigure}

For each word tested in the story, cosine similarity was computed between the GloVe embeddings \cite{pennington2014glove, michelmannMomentbymomentTrackingNaturalistic2021} for the responded word and the correct answer, and averaged across participants. In contrast to a binary scoring approach (correct vs. incorrect), this 
allows partial credit 
to be assigned for semantically similar responses to the correct answer (Appendix \ref{methods}). Replicating the findings in Michelmann et al. \cite{michelmannMomentbymomentTrackingNaturalistic2021}, we found that the story-exposure group significantly outperformed the no-exposure group in guessing the correct words ($p < 0.001$; one-tailed test; Figure \ref{fig:histograms}).

\subsection{Model-based word surprisal is related to human memory for spoken narratives}

\begin{figure}[h]
    \centering
    \includegraphics[width=\textwidth,  trim={0cm 0cm 0cm 0cm}]{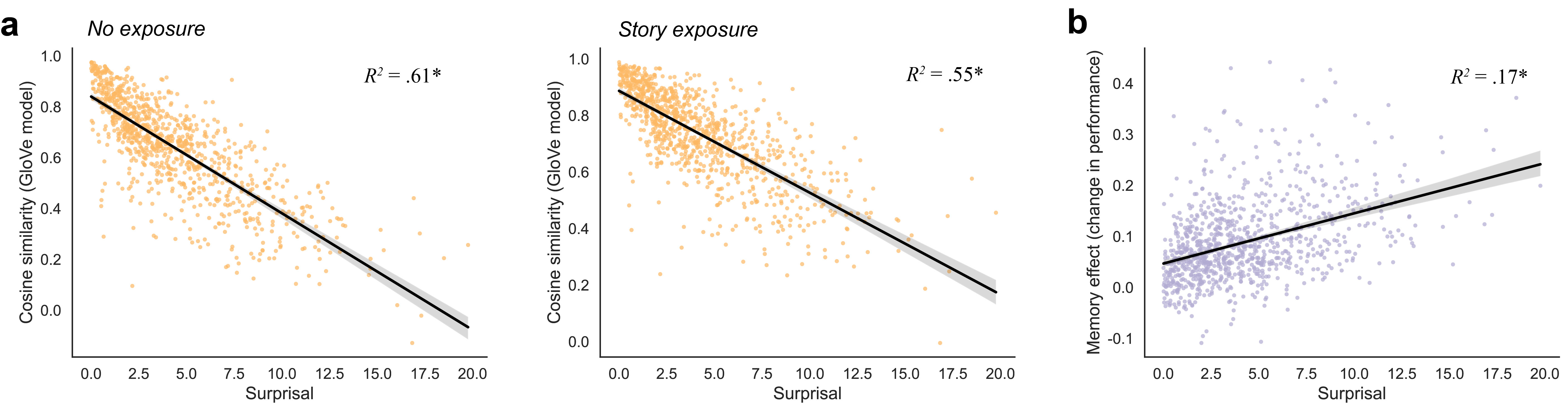}
    \caption{Predicting human behavior with GPT-2 surprise. (A) Model-based surprisal strongly predicts cloze performance, regardless of story exposure. (B) Surprisal is significantly correlated with memory performance for individual words in a spoken story.}
    \label{fig:correlations}
\end{figure}

We next tested the effect of word surprisal on cloze performance. 
We used GPT-2 to estimate surprisal for each of the 1033 story tokens and combined sub-tokens for each word. We found that word surprisal shows a robust inverse correlation in both the no-exposure ($R^2 = 0.61 $; $p < 0.001$) 
and the story-exposure group ($R^2 = 0.55 $; $p < 0.001$; Figure \ref{fig:correlations}A). This 
indicates that surprising words 
predict lower performance in the cloze task, regardless of prior experience with the story.

We find that GPT-2 estimated surprise is positively correlated with the effect of memory on cloze task performance, i.e., the signed difference between the story-exposure and no-exposure groups ($R^2 = 0.17 $, $p < 0.001$), shown in Figure \ref{fig:correlations}B. This effect was consistent for larger models with better perplexity, with only marginal differences in the overall correlation (GPT-2 medium: $R^2 = 0.13 $, $p < 0.001$; large: $R^2 = 0.13 $, $p < 0.001$; XL: $R^2 = 0.11 $, $p < 0.001$).

However, it is possible that other properties that co-vary with surprise explain this memory effect. 
In particular, we evaluate how word written-frequency and distinctiveness affect the relationship between LM-based surprise and memory performance (Appendix Figure \ref{fig:covariance}). Measures of distinctiveness describe the uniqueness of a word's semantic associations, and have been shown to strongly drive lexical memorability \cite{tuckuteIntrinsicallyMemorableWords2018}. We define distinctiveness as the average GloVe dissimilarity between a word and all other words in the story. Note that this represents story-specific distinctiveness, which differs from the measure in \cite{tuckuteIntrinsicallyMemorableWords2018} that quantifies average dissimilarity between a word and all other words in a large corpus. Word frequency, i.e. the unigram probability of a given word, quantifies the overall exposure to a word in a large corpus of text \cite{balota_english_2007}. We fit these features along with word surprisal as predictors in a multiple regression model to evaluate their overall and individual impact on memory performance. Importantly, this allows us to quantify the role of context in memory, while taking into account lexical features such as distinctiveness and word frequency. As expected, we found that the overall model significantly predicts memory performance, with a higher $R^2$ than word surprisal alone ($R^2 = .24$, $p < 0.001$). 
Each normalized $\beta$ coefficient shows a significant contribution in predicting memory (distinctiveness $\beta = 0.011 $, $p < 0.001$; word surprisal $\beta = 0.012 $, $p < 0.001$; written-frequency $\beta = -0.02 $, $p < 0.001$, Appendix Figure \ref{fig:glm_coeffs}). These findings suggest that word surprisal in GPT-2 predicts human memory performance for narratives, indicating its candidacy as an encoding policy. 

To further understand the effect of context, we next examined how varying the context length given to GPT-2 affected its ability to predict memory performance. Because of the power-law scaling of language model performance as a function of context length \cite{kaplanScalingLawsNeural2020}, we selected roughly logarithmically spaced lengths. We found the correlation between surprisal and memory effect improves with longer windows, and plateaus at around 600-token length (Appendix Figure \ref{fig:windows}). The correlations at all context lengths were significant using permutation tests ($p < 0.001$).

\subsection{Testing the ability of attention weights to predict human memory}

Prior work has proposed that attention weights could serve as a memory encoding mechanism \cite{raeCompressiveTransformersLongRange2019}. Extensive research on the role of attention in human memory \cite{chun2007interactions,aly2016attention,aly2017hippocampal} suggests this may be a feasible linking hypothesis, although see \cite{lindsayAttentionPsychologyNeuroscience2020}. In our dataset, we tested whether memory performance was correlated with attention weights from several of the 12 layers, and found either no relation or an extremely weak one for layer 11 (Appendix Figure \ref{fig:attn_regression}).

\section{Discussion and Future Directions}
\label{headings}

In this article, we detailed an approach for augmenting LMs on the basis of the human memory literature. We also provided an empirical demonstration for validating candidate mechanisms through a principled comparison with human behavioral data. In particular, we tested the hypothesis that word surprisal in GPT-2 would predict human memory performance in a narratives dataset \cite{michelmannMomentbymomentTrackingNaturalistic2021}, a prediction borne from the human memory literature more broadly \cite{sinclairSurpriseDestabilizePrediction2018, fosterRoleSurpriseLearning2019, antonyBehavioralPhysiologicalNeural2020, futrellLossyContextSurprisalInformationTheoretic2020, haeuserEffectsPredictionError2021}. We found that surprisal significantly correlated with human memory in this task, indicating its viability as a candidate encoding policy for future architectures. For example, surprisal could determine whether some context or word is written to an external memory. A similar encoding policy has been attempted in lifelong learning setups with mixed success \cite{ramalhoADAPTIVEPOSTERIORLEARNING2019, dautumeEpisodicMemoryLifelong2019}, suggesting that understanding the interaction of surprisal with other variables affecting memory may improve the policy \cite{rouhaniDissociableEffectsSurprising2018}.

Several other limitations should be identified in our empirical approach. First, the story that was used only consisted of 1033 tokens, which only exceeds the input size of GPT-2 by a small margin. Future work would benefit from using texts which significantly exceed 1024 tokens to better address temporal generalization limits in Transformer models. In addition, evaluating model performance on naturalistic free-recall data would provide a more ecologically valid benchmark for comparison \cite{hamiltonRevolutionWillNot2020}. Future work should also consider a more comprehensive evaluation of surprisal-related measures which can be generated using Transformer outputs or parameter states. In this work, we also define two types of memory-augmented Transformers, static and dynamic, which have direct consequences for successful modification. Research in this area would benefit by continuing to develop a taxonomy of model types which integrates explicit memory mechanisms to facilitate further progress.

\bibliography{references}
\bibliographystyle{unsrt}

\newpage

\appendix

\renewcommand\thefigure{\thesection.\arabic{figure}}
\setcounter{figure}{0}

\section{Supplemental Materials and Methods}
\label{methods}

\paragraph{Dataset} 
To evaluate model parameters with respect to human memory performance, we used a previously published behavioral dataset \cite{michelmannMomentbymomentTrackingNaturalistic2021} that was made publicly available through a Creative Commons 4.0 license. This dataset contains data collected online in which an experimental group of participants listened to a spoken story before performing a behavioral task. All participants provided informed consent in accordance with the local Institutional Review Board.

The behavioral experiment was run online (Amazon’s Mechanical MTurk) across two groups of volunteer participants \cite{michelmannMomentbymomentTrackingNaturalistic2021}. In total, 100 participants were collected. We used anonymized data from these participants in the current study (“replication” dataset in \cite{michelmannMomentbymomentTrackingNaturalistic2021}, which is the only shared dataset that provides participant-level results). 

\paragraph{Story materials}
Participants listened to a humorous story which is 7 minutes and 30 seconds in duration and contained 965 words (recorded live as part of the “The Moth” storytelling event in New York City; “Pieman” by Jim O’ Grady). The only offensive content in the story is the use of occasional swear words. The transcript for this story was provided as part of the public dataset release.

\paragraph{Experimental procedures}
All participants took part in a cloze task \cite{taylorClozeProcedureNew1953}, which provided participants with text  from the story and asked them to guess the next word. The cloze task is widely used in the psychological literature to estimate the probability of the next word given some context, making this behavioral design particularly suitable for comparison with probabilistic predictions computed from language models. In the current paradigm, participants saw 10 words from the transcribed story and asked to guess the word that followed (continuing for every word in the story). The task starts with the third word in the story, limiting the context for words in the beginning of the story \cite{michelmannMomentbymomentTrackingNaturalistic2021}. In order to capture memory for the story, participants were split into two experimental groups. In the first group (N = 50), participants listened to the story once before performing the cloze task (“story-exposure group”), while the second group (N = 50) did not listen to the story before performing the cloze task (“no-exposure group”). The signed difference in performance between the story-exposure group and the no-exposure group represented the improvement in performance as a function of recalling the story from memory. The paradigm and experimental procedures are described in greater detail in \cite{michelmannMomentbymomentTrackingNaturalistic2021}.

\paragraph{Behavioral data analysis} 
As acquired from \cite{michelmannMomentbymomentTrackingNaturalistic2021}, cloze task performance was computed using GloVe vector embeddings (pre-trained on Common Crawl data \cite{pennington2014glove}). Specifically, participants’ word predictions were scored by computing cosine-similarity between the vector embeddings for the predicted word and the correct word. The average semantic similarity among participants for a particular word represents the group-level performance for that word. We compared these word similarity scores across the story-exposure and no-exposure groups to evaluate differences in performance as a function of story recollection. Notably, this approach provides a more continuous measure of performance as opposed to a binary dependent measure (correct or incorrect) as used in the main text of Michelmann et al. (\cite{michelmannMomentbymomentTrackingNaturalistic2021}; but see Michelmann Supplementary Method 2 and Supplementary Note 3). In other words, this measure of performance can be thought to provide a more sensitive measure than binary scoring, in that some credit can be assigned in the case that a participant predicts a semantically similar, but not identical, word to the correct answer.

\paragraph{Word surprisal as negative-log likelihood}
We computed word surprisal for each word in the story using GPT-2 \cite{radfordLanguageModelsAre}, with the GPT-2 tokenizer and model from Huggingface \cite{wolf2019huggingface}. For each of the 1033 sub-word tokens in the story, we included all previous tokens as preceding context (up to the maximum 1024). This ensured that we maximized contextual information for each token. We then take the cross entropy loss for the last token.

The surprisal for each word in the story is represented by the negative log likelihood generated by the loss function:

\[S_{model}\ (x_i\ )=-logP_{model}\ (x_i\ |x_{j<i})\]

To compute the overall negative log likelihood of a word, we simply sum the negative log likelihood values of the component sub-tokens. As such, a higher value represents a more surprising word given its context according to the model. We refer to the inverse log likelihood at the word-level as word surprisal.

\paragraph{Factoring in word frequency and distinctiveness}
Past research has shown that both word frequency \cite{lohnas_parametric_2013} and distinctiveness \cite{tuckuteIntrinsicallyMemorableWords2018} have a significant impact on word retention. We sought to understand to what extent these factors along with word surprisal predict memory performance in the current dataset. We acquired the word written-frequency for each word in the story using the Hyperspace Analogue to Language (HAL) frequency norms \cite{balota_english_2007}. Word-distinctiveness, or how few unique word associations a word possesses, was computed using GloVe word embeddings \cite{penningtonGloveGlobalVectors2014}. In particular, we computed the cosine-similarity between each word in our story and all of the other words in the story. Taking one minus the average of these similarity values provided us with a story-level distinctiveness value for each word, such that higher values represent more distinctive words.
We fit word written-frequency and distinctiveness along with word surprisal as predictors in a general linear model (GLM) to predict memory performance. Notably, words for which written-frequency was not available \cite{balota_english_2007}, or for which we were not able to compute distinctiveness, were excluded from this analysis (19 words excluded in total)

\paragraph{Statistical testing}
Throughout this work, statistical analysis was applied using nonparametric permutation tests. Across comparisons, we implemented 10,000 permutations to ensure a reliable estimate of the null distribution. Significance was evaluated at $p < 0.05$. Note, we indicate the use of a one-tailed test when an effect is evaluated in a specific direction, otherwise a two-tailed statistic is reported. 

\newpage
\section{Supplemental Results}
\label{supp_results}

\begin{figure}[!h]
    \centering
    \includegraphics[width=\textwidth, trim={0 0 0 0}]{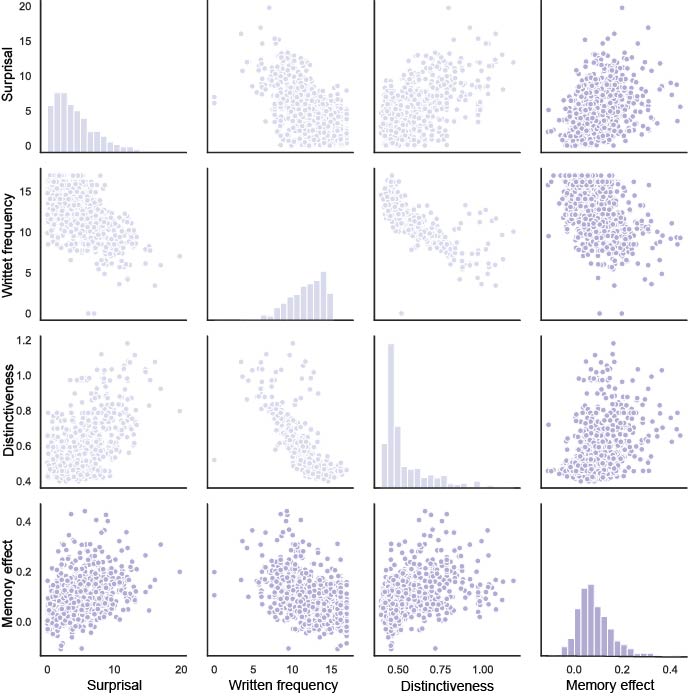}
    \caption{Relationships across predictors and memory effect (cosine similarity) in multiple regression analysis.}
    \label{fig:covariance}
\end{figure}

\begin{figure}[!h]
\begin{minipage}[t]{0.5\linewidth}
    \centering
    \includegraphics[width=\textwidth]{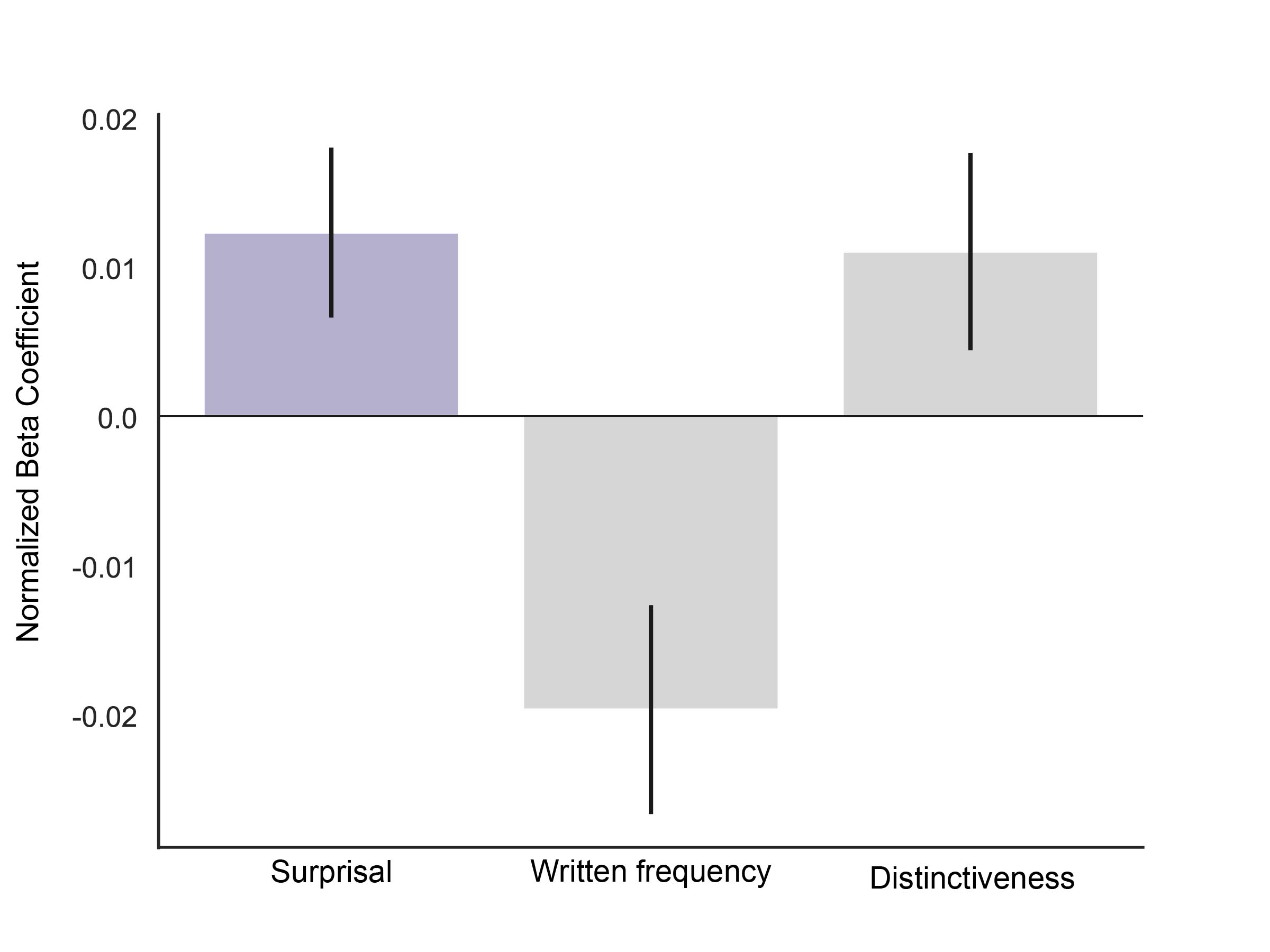}
    \caption{Normalized beta coefficients for predictors in multiple regression analysis. Error bars represent 95\% bootstrapped confidence intervals (10,000 bootstrap iterations).}
    \label{fig:glm_coeffs}
\end{minipage}%
\end{figure}
    
\begin{figure}[!h]
\begin{minipage}[t]{0.5\linewidth}
    \centering
    \includegraphics[width=\textwidth]{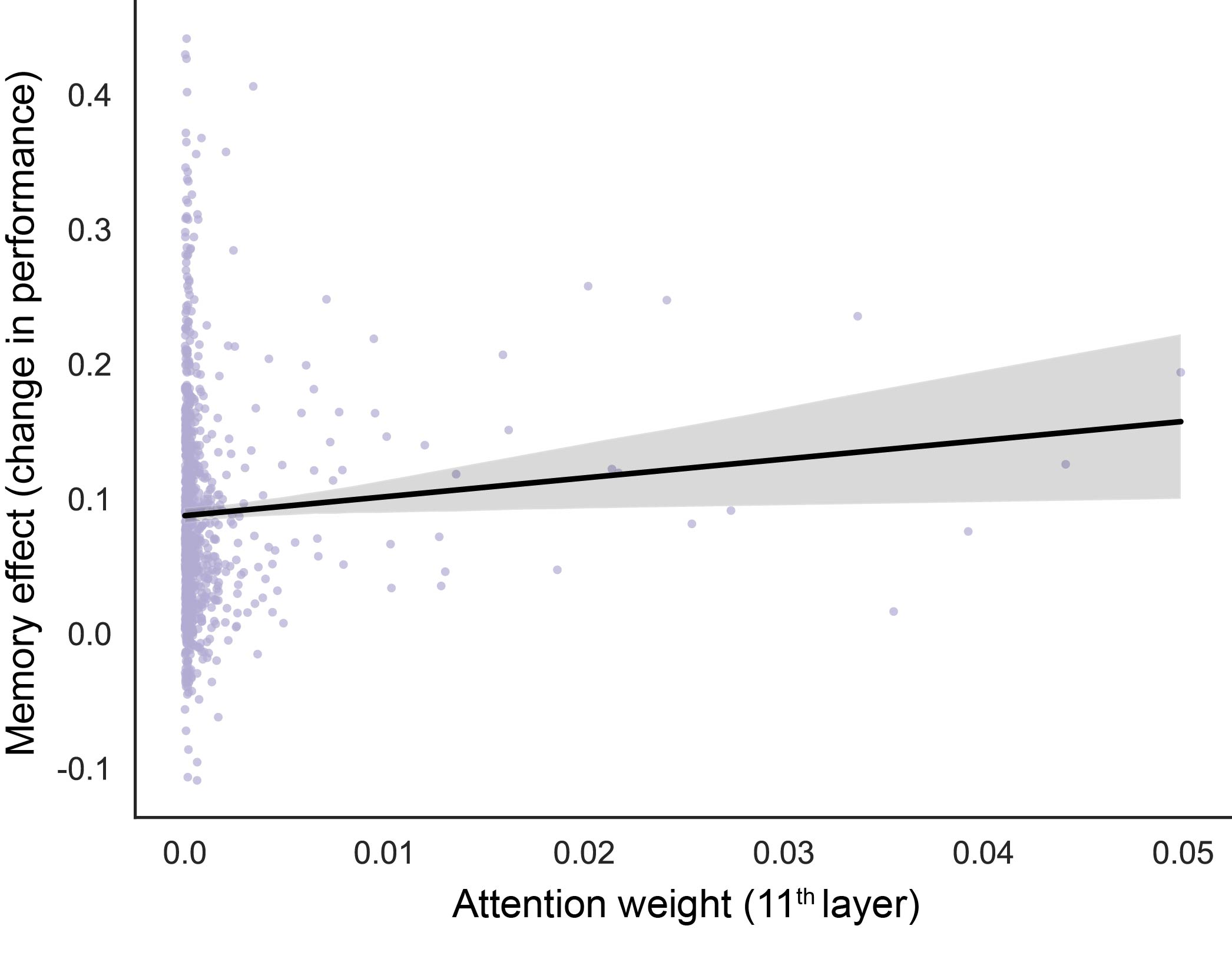}
    \caption{Attention weights for 11\textsuperscript{th} layer in GPT-2 versus memory effect ($R^2 = 005 $; $p = 0.036$). We did not find a significant effect in the 1\textsuperscript{st}, 6\textsuperscript{th}, and 12\textsuperscript{th} layers ($p > 0.05$).}
    \label{fig:attn_regression}
\end{minipage} 
\end{figure}

\begin{figure}[!h]
\begin{minipage}[t]{0.5\linewidth}
    \centering
    \includegraphics[width=\textwidth]{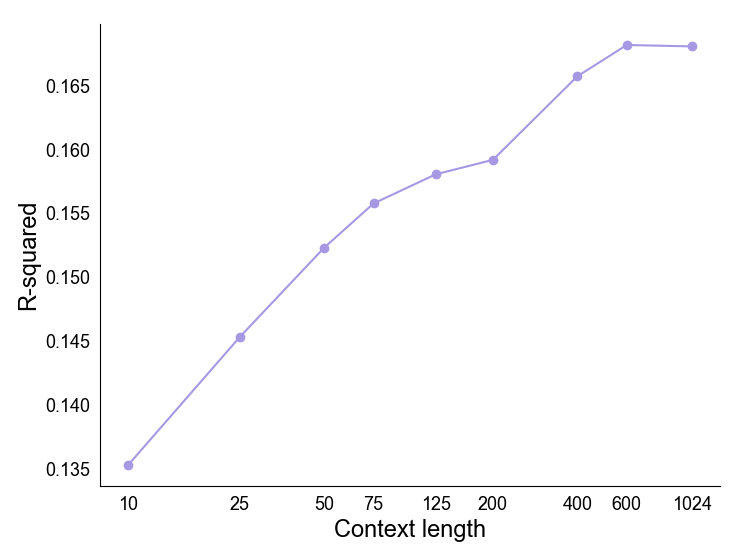}
    \caption{R-squared for Pearson correlation between the memory effect and GPT-2 surprisal as a function of input window sizes. Permutation tests indicate all correlations are significant ($p < 0.001$).}
    \label{fig:windows}
\end{minipage}%
\end{figure}
    
\end{document}